\documentclass[runningheads]{llncs}
\usepackage{graphicx}

\usepackage{wrapfig,lipsum,booktabs}
\usepackage{tikz}
\usepackage{comment}
\usepackage{amsmath,amssymb}
\usepackage{color}
\usepackage{graphicx}
\usepackage{amsmath}
\usepackage{amssymb}
\usepackage{booktabs}
\usepackage{shtabularlines}
\usepackage[normalem]{ulem}
\useunder{\uline}{\ul}{}
\usepackage{pifont}
\usepackage{multirow}
\usepackage[pagebackref,breaklinks,colorlinks]{hyperref}

\newcommand*{\Scale}[2][4]{\scalebox{#1}{$#2$}}%

\newcommand\blfootnote[1]{%
  \begingroup
  \renewcommand\thefootnote{}\footnote{#1}%
  \addtocounter{footnote}{-1}%
  \endgroup
}

\usepackage[accsupp]{axessibility}

\begin{document}
\pagestyle{headings}
\mainmatter
\title{End-to-End Instance Edge Detection}

\titlerunning{}

\author{Xueyan Zou$^*$ \and
Haotian Liu$^\dagger$ \and
Yong Jae Lee }
\authorrunning{X. Zou et al.}
\institute{University of Wisconsin--Madison\\
\email{\{xueyan,lht,yongjaelee\}@cs.wisc.edu}
\vspace{6pt}
}

\maketitle
\begin{center}
    \centering
    \includegraphics[width=1.\textwidth]{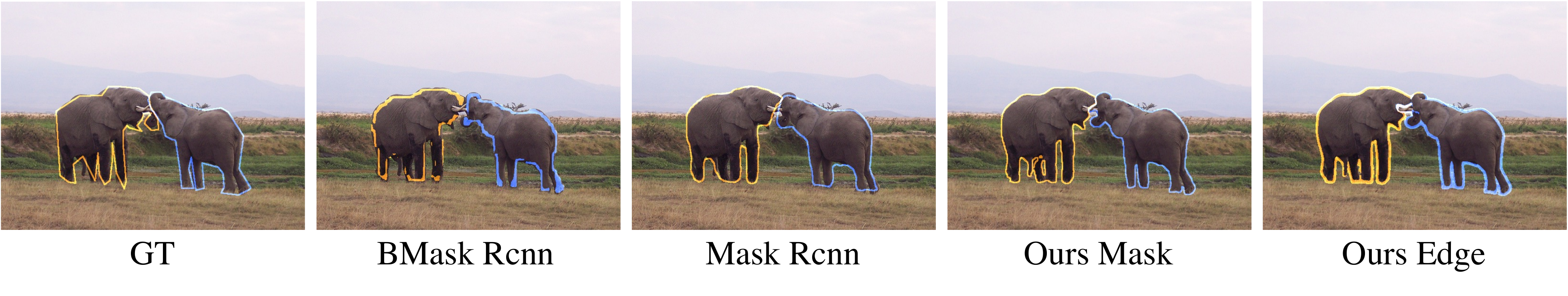}
    \vspace{-20pt}
\end{center}

\begin{abstract}
Edge detection has long been an important problem in the field of computer vision. Previous works have explored category-agnostic or category-aware edge detection. In this paper, we explore edge detection in the context of object \emph{instances}.  
Although object boundaries could be easily derived from segmentation masks, in practice, instance segmentation models are trained to maximize IoU to the ground-truth mask, which means that segmentation boundaries are not enforced to precisely align with ground-truth edge boundaries.
Thus, the task of instance edge detection itself is different and critical.
Since precise edge detection requires high resolution feature maps, we design a novel transformer architecture that efficiently combines a FPN and a transformer decoder to enable cross attention on multi-scale high resolution feature maps within a reasonable computation budget.
Further, we propose a light weight dense prediction head that is applicable to both instance edge and mask detection.
Finally, we use a penalty reduced focal loss to effectively train the model with point supervision on instance edges, which can reduce annotation costs.
We demonstrate highly competitive instance edge detection performance compared to state-of-the-art baselines, and also show that the proposed task and loss are complementary to instance segmentation and object detection.
\end{abstract}

\section{Introduction}
\blfootnote{$^{*,\dagger}$ Work done during an internship at Cruise.}
We address the problem of instance edge detection. Unlike category-agnostic \cite{martin2004learning,arbelaez2010contour,xie2015holistically} or category-aware (semantic) edge detection \cite{yu2017casenet,yu2018simultaneous}, instance edge detection requires predicting the semantic edge boundaries of \emph{each object instance}. This problem is fundamental and can be of great importance to a variety of computer vision tasks including segmentation, detection/recognition, tracking and motion analysis. In particular, instance edge detection can be critical for applications that require precise object boundary localization such as autonomous driving or robot grasping.

\begin{figure*}[t]
  \centering
  \includegraphics[width=1.\textwidth]{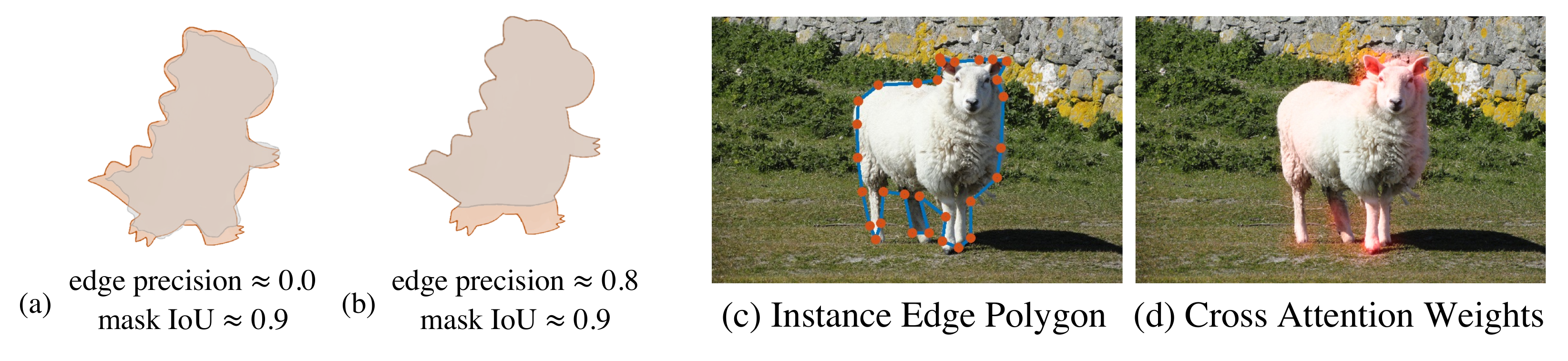}
  \caption{(a,b) Although closely related, instance segmentation and edge detection are not fully invertible; i.e., a method that performs well on instance segmentation will not necessarily perform well on edge detection. In this example, although the mask IoUs in (a) and (b) are the same, their edge precision are very different. (c)  Although the annotated points (red) are on the object's boundary, the edges (blue) that connect them do not align well to the object’s boundary. (d) Cross attention weights between object queries and image feature in DETR~\cite{carion2020end}.} 
  \label{fig:moti}
\end{figure*}

Instance segmentation is closely related to instance edge detection.  After all, in theory, an instance's boundary can be trivially extracted from the output of any standard instance segmentation algorithm~\cite{he2017mask,bolya2019yolact}.  However, in practice, this naive solution does not produce good results~\cite{cheng2020boundary,cheng2021boundary}. Since an instance segmentation algorithm is trained to correctly predict \emph{all} pixels that belong to an object, and since there are relatively few pixels on an instance's contour than inside of it, the model has no strong incentive to accurately localize the instance boundaries. As shown in Fig.~\ref{fig:moti} (a), although the predicted mask may have high quality pixel alignment with the ground truth mask, its boundary may not be well-aligned with the ground truth instance edge. Thus, instance edge detection itself is a unique and important task.

Meanwhile, the recent transformer~\cite{vaswani2017attention} based DETR~\cite{carion2020end} object detector has drawn significant attention as it greatly simplifies the detection pipeline by achieving end-to-end learning without ROI pooling, NMS, and anchor modules. Moreover, several transformer based object detection models~\cite{carion2020end,meng2021conditional,chen2021pix2seq} have shown that object boundaries produce high responses in the attention maps (Fig.~\ref{fig:moti} (d)), which suggests that transformer based architectures can be suitable for instance edge detection. However, DETR is not directly applicable to instance edge detection in several ways. First, the quadratic complexity on self-attention over feature maps prevents DETR from using high resolution feature maps in its transformer decoder. However, high resolution features maps are critical for dense prediction tasks such as edge detection. Second, since the output dimension (length) of edges can vary per object instance (unlike box coordinates or classes), it is not straightforward to produce instance edge outputs directly from an object query. 

To address these difficulties, we first propose a multi-scale transformer decoder that takes in both encoder features and FPN features. The first several layers in the transformer decoder take in the feature maps computed from the transformer encoder while the last two layers take in the high resolution FPN feature maps. In this way, object queries interact with different resolution feature maps in a coarse to fine grained manner. To enable the model to output instance edge detections, we  introduce a light weight dense prediction head that computes a simple matrix multiplication between object queries and high resolution feature maps to produce binary output maps (whose spatial resolution is the same as the feature maps) where predicted 1/0's indicate edges/non-edges.

By changing only the loss function, we show that our method can perform either edge detection or instance segmentation without any modification to the architecture.  At the same time, since instance edge detection and segmentation are closely related, if we do perform both tasks together (with separate heads for each task), we show that they provide complementary benefits to each other.

Finally, one key challenge with instance edge detection is its annotation requirement; i.e., labeling all pixels along an object instance's contour can be extremely expensive. We therefore propose to train our instance edge detector using only \emph{point supervision}.  Similar to how instance segmentation methods are trained with keypoint-based polygon masks \cite{castrejon2017annotating,acuna2018efficient}, we use a sparse set of keypoint annotations along the object's boundary. For instance segmentation, this results in a 4.7x speed up over annotating all points~\cite{castrejon2017annotating}.  However, due to the sparsity, simply connecting adjacent keypoints to `complete the edge' (as done in BMask R-CNN~\cite{cheng2020boundary}) can often lead to incorrect annotations, as shown in Fig.~\ref{fig:moti} (c). We therefore instead train the edge detector using the keypoints with a penalty reduced loss along the edges.

\vspace{-5pt}
\subsubsection{Contributions.} (1) We introduce a novel transformer model with a multi-scale transformer decoder and dense prediction head for instance edge detection, which achieves highly competitive results on the COCO and LVIS datasets compared to related state-of-the-art baselines.  (2) We demonstrate that our model can perform object detection, instance segmentation, and instance edge detection in a single pass, and show complementary benefits for each task. (3) We show that we can efficiently train our instance edge detection model with only point supervision using a penalty reduced focal loss.

\section{Related Work}

\subsubsection{Edge Detection.} 
Edge detection has a rich history and has been studied since at least the 1980s. Early pioneering methods include the Sobel filter~\cite{kittler1983accuracy}, zero-crossing~\cite{marr1980theory,torre1986edge}, and Canny edge detector~\cite{canny1986computational}. The early 2000s saw approaches driven by information theory such as statistical edges~\cite{konishi2003statistical}, Pb~\cite{martin2004learning} and gPB~\cite{arbelaez2010contour}. The advent of deep learning in the last decade introduced highly effective approaches like HED~\cite{xie2015holistically,liu2017richer}, and the focus shifted from texture based edge detection~\cite{arbelaez2010contour} to category agnostic semantic edge detection~\cite{hariharan2014simultaneous}. More recently, researchers have begun to focus on \emph{semantic aware} edge detection~\cite{yu2017casenet,yu2018simultaneous} whose goal is to accurately localize the boundary between semantic classes (but not between instances). In this work, we explore edge detection in the \emph{semantic and instance aware} setting~\cite{cheng2020boundary} to localize object \emph{instance} boundaries.

\vspace{-5pt}
\subsubsection{Instance Segmentation.}  
Instance segmentation is now a classic task in computer vision, where early methods~\cite{girshick2014rich,hariharan2014simultaneous,dai2015convolutional,hariharan2015hypercolumns} resorted to classifying bottom-up segments. The development of the RCNN framework~\cite{he2015spatial,girshick2015fast,ren2015faster} led to Mask R-CNN~\cite{he2017mask}, a strong performing and simple architecture, which greatly increased the popularity of instance segmentation.  Since then, researchers have explored various directions to improve efficiency~\cite{bolya2019yolact,peng2020deep} and effectiveness~\cite{huang2019mask,kirillov2020pointrend,cheng2020boundary}. Recent trends are towards developing light weight detectors that  contain only one-stage~\cite{bolya2019yolact}, without anchors~\cite{tian2020conditional}, ROI-align and NMS~\cite{hu2021istr}. Further, instead of formulating instance segmentation as a dense pixel prediction task~\cite{long2015fully,he2017mask}, some approaches~\cite{castrejon2017annotating,liang2020polytransform} predict polygon points for each instance to focus more on the object boundaries. Another uses a polar representation~\cite{xie2020polarmask}.

The recent concurrent works, MaskFormer~\cite{cheng2021per} and Mask2Former~\cite{cheng2021masked}, are similar to our work in regards to the dense prediction head and pixel decoder.  However, our main focus is on solving the different instance edge detection task (while also performing bounding box detection and instance segmentation), whereas those works focus on segmentation. Importantly, we find that simply changing their models to perform bounding box and mask prediction in the transformer decoder results in low detection accuracy, suggesting that sophisticated changes to the methods would be needed to do well on those tasks.

Finally, related to our point supervised edge detection loss, \cite{kirillov2020pointrend,cheng2021pointly} also perform point supervised learning, however, they sample points on the model's predicted feature maps instead of using ground truth annotations. Similarly, although \cite{li2021fully} performs sparse sampling on output mask, it is designed for the panoptic segmentation task, which is different from our goal of instance edge detection.

\vspace{-5pt}
\subsubsection{Transformers.}  The Transformer was first introduced in~\cite{vaswani2017attention}, and has become the state-of-the-art architecture for natural language processing tasks~\cite{tenney2019bert,zaheer2020big}. However, despite its high accuracy, the transformer architecture suffers from slow convergence~\cite{liu2020understanding} and quadratic computation and memory consumption~\cite{kitaev2020reformer,beltagy2020longformer} necessitating a high number of GPUs and up to weeks for training. Recently, the transformer has begun to be explored for visual recognition tasks including image classification~\cite{dosovitskiy2020image}, detection~\cite{carion2020end}, image generation~\cite{jiang2021transgan}, etc. Since image data typically has longer input sequences (pixels) than text data, the computation and memory problem is arguably more critical in this setting. To address this, researchers have proposed methods \cite{zhang2021multi,wang2021pyramid,liu2021swin} that reduce both computation and memory complexity, allowing the transformer to perform dense prediction tasks~\cite{xie2021segformer,cheng2021per,strudel2021segmenter,wang2021max}. Apart from the efficiency problem, the vision transformer also suffers from long training times especially for object detection; specifically, 500 epochs for DETR~\cite{carion2020end} to achieve the same performance as Faster R-CNN~\cite{ren2015faster} with only 50 epochs. As such, many follow-up works to DETR aim at improving  its convergence speed via additional priors~\cite{zhu2020deformable,gao2021fast,meng2021conditional} or by reducing transformer density~\cite{zhu2020deformable}. In particular,~\cite{meng2021conditional} achieves significant improvements by introducing a conditional spatial query.  In this work, we extend the DETR framework~\cite{meng2021conditional,carion2020end} to instance edge detection and segmentation.

\section{Method}
\subsection{Problem Definition}

Given an input image $I$, the task of instance edge detection is to correctly predict the boundaries of each object instance $G_E=\{e_0, e_1, ..., e_n\}$ together with its category label $G_C=\{l_0, l_1, ..., l_n\}$, where $n$ is the number of instances in $I$.

\subsection{Method Overview}
\begin{figure*}[t]
  \centering
  \includegraphics[width=1.\textwidth]{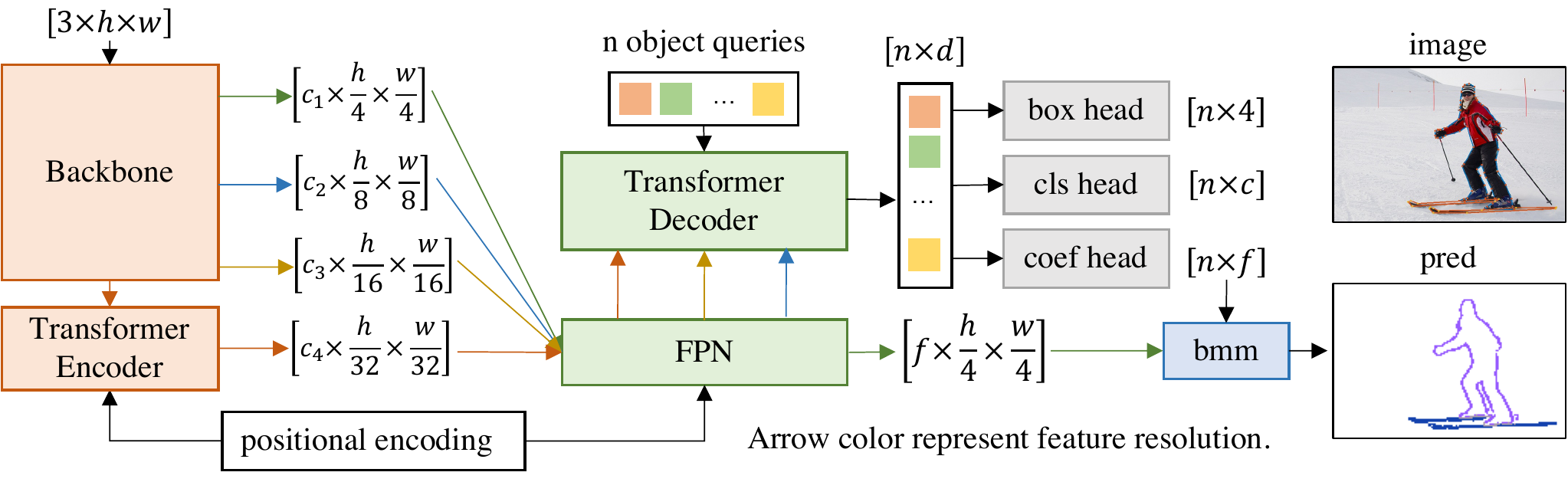}
  \caption{\textbf{Method overview.} Our model consists of four main components: a backbone feature extractor, FPN, multi-scale transformer decoder, and task heads. The output dimension for each module is indicated in $[\cdot]$, and bmm denotes batch matrix multiplication.}
  \label{fig:method}
\end{figure*}

Our method overview is shown in Fig.~\ref{fig:method}. It comprises four main components: (1) a backbone network, which extracts a hierarchical combination of features (Sec.~\ref{sec:backbone}) and a transformer encoder network that perform global attention over low resolution feature maps; (2) a feature pyramid network (FPN)~\cite{lin2017feature}, which fuses the feature maps of different levels (Sec.~\ref{sec:fpn}); (3) a multi-scale transformer decoder, which takes in the different resolution feature maps and a set of object queries and performs cross-attention between them (Sec.~\ref{sec:decoder}); and (4) light weight dense prediction heads, which perform either instance edge detection or segmentation (Sec.~\ref{sec:dense}), along with prediction heads for box detection and classification. We also introduce our point based instance edge detection loss in Sec.~\ref{sec:pointedge} and analyze relationship between object edge and segmentation in Sec.~\ref{sec:relation}. \textit{The multi-scale transformer decoder, light weight dense prediction head, point based instance edge detection loss, and object edge and segmentation relationship analysis are the main technical contributions.} 

\vspace{-5pt}
\subsubsection{Backbone Feature Extractor}\label{sec:backbone}
Given an input image $I$ with shape $[3,h,w]$, the feature backbone extracts a set of feature maps with shape $[c_i,h/r_i,w/r_i]$ for $c_i \in [256,512,1024,2048]$ and $r_i \in [4,8,16,32]$.  We set the feature extractor to be a ResNet~\cite{he2016deep} together with a transformer encoder~\cite{carion2020end} with self-attention~\cite{vaswani2017attention}.

\vspace{-5pt}
\subsubsection{Feature Pyramid Network}\label{sec:fpn}
The final output of the transformer encoder is $1/32$ of the original image, which is too low for edge detection. Thus, we integrate a feature pyramid network (FPN)~\cite{lin2017feature} to increase the resolution by fusing higher resolution feature maps from the backbone and self-attention features. The output of the FPN includes feature maps with $[1/4,1/8,1/16,1/32]$ of the original image resolution. We also add positional encodings~\cite{vaswani2017attention} to the projected features, which will enable the object queries (explained shortly in Sec.~\ref{sec:decoder}) to better localize objects and their boundaries.

\vspace{-5pt}
\subsubsection{Transformer Decoder}\label{sec:decoder}
Given $n$ input object queries each with $d$ dimensions (i.e., size $[n,d]$), the transformer decoder first applies self-attention~\cite{vaswani2017attention} to allow the object queries to interact with each other to remove redundant predictions. It then applies cross attention~\cite{vaswani2017attention} between the object queries $Q$ with shape $[n,d]$ and the feature map $F$ (i.e., size $[d,h/i,w/i]$ for $i\in (32,16,8)$ with $(h,w)$ as the image size) from the FPN. Note that the cross attention operation has a time and memory complexity of $O(nd(hw)^2 + nd^2(hw))$. Thus, in order to leverage high resolution feature maps without a large computation overhead, we use a feature resolution of $1/32$ in the first four layers and $1/16$ and $1/8$ in the last two layers of transformer decoder. In this way, the object queries can attend to the features in a coarse to fine-grained manner for improved dense prediction performance.

\vspace{-5pt}
\subsubsection{Dense Prediction Head}\label{sec:dense}
Our design for edge prediction is motivated by three observations: (1) Without training with any dense pixel-level labels, and instead, with only box supervision, the cross attention maps computed between the object queries and encoder features have the nature to focus on instance edges~\cite{carion2020end,meng2021conditional,chen2021pix2seq} (Fig.~\ref{fig:moti} (d)). (2) The encoder features within the same object \emph{instance} have similar representations~\cite{carion2020end}. (3) By directly taking a weighted combination of the high-resolution feature maps along channel dimension, it leads to mask predictions that can clearly follow the boundaries of the instances~\cite{bolya2019yolact}. These three observations suggest that convolving the feature maps with each object query could lead to accurate pixel-level instance edge predictions.

Given the transformer decoded object queries $Q$ with shape $[n, d]$, and image features $F$ from the FPN with shape $[f,h/4,w/4]$, we first predict $f$ weight coefficients for each object query with a simple linear projection:
\begin{gather}
    Q' = sigmoid(linear(d,f)(Q))
\end{gather}
where $linear(i,j)$ indicates linear projection from dimension $i$ to $j$. The result is a coefficient for each query; i.e., coefficient tensor with shape $[n, f]$.  This operation corresponds to the `coef head' shown in Fig.~\ref{fig:method}.

Then, to predict the edge map for each object query, we apply a $1 \times 1$ convolution to the feature maps $F$ using the object query $Q'$ coefficients as filter weights. This is equivalent to applying a batch matrix multiplication between $Q'$ and $F$:
\begin{equation}
    O_i = sigmoid(Q'_i \times F), \forall i
\end{equation}
where $i$ is the index of the object query, $Q'_i$ has shape $[1,f]$, $F$ has shape $[f,h,w]$, and $O_i$ has shape $[h,w]$. Note that all object queries are multiplied with the same set of features maps. This dense prediction head is general and very light weight, and is applicable to any object instance based pixel classification tasks. For example, we can easily obtain mask segmentations by only changing the edge detection loss function to a mask segmentation loss.

\begin{figure}[t]
   \centering
   \includegraphics[width=1.0\textwidth]{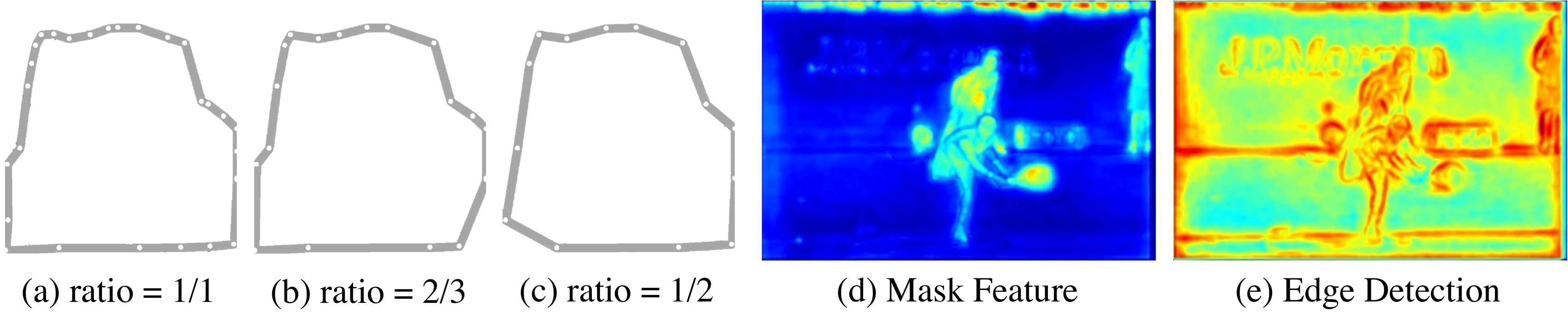}
   \caption{ (a,b,c) show the annotated points that are regarded as positive samples and the tunnels (in gray) that are inside the penalty reduced regions. We show these with different fractions of annotated points. (d,e) Feature map norm on channel dimension for the layer preceding the final output layer (pixel-wise mask/edge for instance/edge detection).}
   \label{fig:method_fig}
\end{figure}

\subsection{Point Supervised Focal Loss}\label{sec:pointedge}
As dense labeling of all pixels along an object instance's contour can be extremely expensive, we train our instance edge detector using only point supervision along the object's boundary, similar to how instance segmentation methods are trained with keypoint-based polygon masks~\cite{castrejon2017annotating,acuna2018efficient}.  Note that simply connecting adjacent keypoints to `complete the edge' as done in BMask R-CNN~\cite{cheng2020boundary} will lead to incorrect annotations that are not on the ground-truth edge (see Fig.~\ref{fig:moti} (c)).

To address this, we design a novel training objective to account for the sparse keypoint annotation. Specifically, we build upon the penalty-reduced pixel-wise logistic regression with focal loss~\cite{law2018cornernet}, which was designed to reduce the penalty in slightly mis-predicted corners of a bounding box (since those slightly shifted boxes will also localize the object well).  In our case, we can use this loss to account for slightly mispredicted keypoints, but we also need to deal with a different issue, which is that a large portion of the ground-truth edges are not annotated at all.  To handle the latter, we construct the ground-truth in the following way.

We first connect the ground-truth keypoints to create edges, and then blur the result with a small $3 \times 3$ kernel (e.g., a Gaussian or a box filter).  This creates a ``tunnel'' whose values are greater than 0.  We set these values to 0.7, and the original keypoints as 1, as shown in Fig.~\ref{fig:method_fig} (a-c).  The lower values for the tunnels account for the uncertainty in ground-truth edge location for the non-keypoints.  While we could also take continuous values that degrade as a function of distance to the keypoints and edges, we find this simple approach to work well in practice.  

Formally, we use the ground-truth maps $Y$ as targets in our extension of the penalty-reduced pixel-wise logistic regression with focal loss~\cite{law2018cornernet}:
\begin{gather}
L_{k} =  \frac{-1}{N}  \sum_{cxy}
\left\{\begin{matrix}
Y_{cxy}(1 -  \widehat{Y}_{cxy})^ \alpha \log(\widehat{Y}_{cxy}) \ \ $if$ \ Y_{cxy} \geq \gamma
\\
(1-Y_{cxy})^\beta(\widehat{Y}_{cxy})^\alpha \log(1-\widehat{Y}_{cxy}) \ \ $else$
\end{matrix}\right.
\end{gather}
where $\alpha$ and $\beta$ are hyper-parameters of focal loss~\cite{lin2017focal}, and $N$ is the number of annotated keypoints inside an image. We set $\alpha=2$ and $\beta=4$ following~\cite{law2018cornernet} and set $\gamma=0.7$. $\widehat{Y}_{cxy}$ and $Y_{cxy}$ denotes the prediction and ground truth value at location $c,x,y$. With this loss, the model is encouraged to accurately predict the annotated edge points, while also predicting edge points inside the `tunnels' that connect those keypoints.
To complement the point supervised focal loss and to get sharper boundaries~\cite{cheng2020boundary}, we also add the dice loss~\cite{milletari2016v} for edge detection.

\subsection{Overall Objective}\label{sec:finalloss}
Our final objective combines the following: for edge detection, we use our point supervised focal loss as well as dice loss~\cite{milletari2016v} between the matched prediction and ground truth edge pairs.  For bounding box regression, we apply L1 and generalized IoU loss~\cite{rezatofighi2019generalized}.  For classification, and to match each object query to a ground truth box, we use the paired matching loss from DETR~\cite{carion2020end}. Finally, when generalizing our architecture to instance segmentation, we follow~\cite{tian2020conditional}, and use the dice loss~\cite{milletari2016v} and sigmoid focal loss.

\subsection{Relation to Instance Segmentation}\label{sec:relation}

Instance edge detection and instance segmentation are highly correlated tasks as their ground truths are fully invertible. However, since the ratio of pixels on the boundary over the inner pixels for an instance mask is very small, an instance segmentation model will have less preference to correctly predict the edge boundaries compared to the inner pixels. In contrast, an edge detection model would fully focus on correctly predicting the edge boundaries as the inner pixels would be labeled as background.

In this section, we provide a quantitative analysis on this difference.  In particular, we perform the analysis using the dice loss~\cite{milletari2016v}, but the conclusion holds for other losses as well. We choose the dice loss because it is used in many state-of-the-art instance segmentation methods~\cite{tian2020conditional,cheng2021per,cheng2021masked} due to its explicit accounting of the imbalance in foreground versus background pixels, which largely improves segmentation performance. 

Given predicted (either edge or mask) instance map $p$ and ground truth map $y$, each with shape $[h,w]$, the dice loss $L(p,y)$ is:
\begin{equation}
\Scale[0.9]{L(p,y) = 1 - (2\textstyle \sum_{j=1}^{hw}p_j y_j)/(\textstyle \sum_{j=1}^{hw}p_j^2 + \sum_{j=1}^{hw}y_j^2)}    
\end{equation}

Its partial derivative with respect to a prediction $p_i$ at pixel $i$ is:
\begin{equation}
 \Scale[0.9]{\frac{\partial L(p,y))}{\partial p_i} = -2(y_i(\textstyle \sum_{j=1}^{hw}p_j^2 + \textstyle \sum_{j=1}^{hw}y_j^2)-2p_i \textstyle \sum_{j=1}^{hw}p_j y_j) / (\textstyle \sum_{j=1}^{hw}p_j^2 + \textstyle \sum_{j=1}^{hw}y_j^2)^2} \label{eq:2}  
\end{equation}

Next, let us consider the case in which pixel $i$ is on the instance boundary ($y_i=1$) but the model incorrectly predicts it as background ($p_i=0$). We would like to analyze the impact of such incorrect boundary predictions when the objective is mask segmentation versus edge detection.  Since a neural network's weights are updated according to their gradient direction and magnitude, the absolute gradient value of a prediction on a single pixel can measure how much it influences the training procedure (given same loss function and prediction value).  We can therefore take the ratio between the absolute gradient value of the boundary pixel's prediction when the objective is edge detection (with predictions and ground truth denoted as $p'_i$ and $y'_i$, respectively) over that when the objective is mask segmentation:
\begin{equation}
\Scale[0.9]{|\frac{\partial L(p',y'))}{\partial p_i'}| / |\frac{\partial L(p,y))}{\partial p_i}| = (\textstyle \sum_{j=1}^{hw}p_j^2 + \textstyle \sum_{j=1}^{hw}y_j^2) /(\textstyle \sum_{j=1}^{hw}p_j'^2 + \textstyle \sum_{j=1}^{hw}y_j'^2) = \alpha \gg 1}  \label{eq:3} 
\end{equation}

For the same object instance, since its mask will be at least as big as its boundary (and typically much larger), we will have $\sum_{j=1}^{hw}y_j^2 \gg \sum_{j=1}^{hw}y_j'^2$, and also $\sum_{j=1}^{hw}p_j^2 \gg \sum_{j=1}^{hw}p_j'^2$ for any reasonably performing model, where $y',p'$ denote edge ground truth and prediction maps, and $y,p$ denote mask ground truth and prediction maps. Thus, we can easily conclude that the gradient magnitude of edge detection will be much larger than that of instance segmentation for pixel predictions on the object boundary, as shown on the right hand side of Eq.~\ref{eq:3}.

In other words, for pixels on the object boundary, the instance edge detection objective enforces a stronger influence than the instance mask segmentation objective. In addition, if we think this about the problem more intuitively, edge detection feature maps will have high response only on the boundary pixels for each object (Fig.\ref{fig:method_fig} (e)) whereas instance segmentation will train towards predicting all pixels within each object (Fig.\ref{fig:method_fig} (d)). And since the ratio of pixels on the boundary over the inner pixels is very small, the model will have less preference to correctly predict the edge boundaries than the inner pixels in mask segmentation.

Thus, we argue that instance edge detection itself is an important task to explore, distinct from instance segmentation, especially for applications that require precise object boundary localization e.g., self driving or robot grasping.

\section{Experiments}
In this section, we first explain the datasets and evaluation metrics used for evaluating instance edge detection. We then present our implementation details. We further describe our key baselines, and compare to them both quantitatively and qualitatively. Finally, we ablate our model with various baseline components. 

\vspace{-10pt}
\subsubsection{Datasets.}
We train our model on MS COCO~\cite{lin2014microsoft} and evaluate on both COCO as well as LVIS~\cite{gupta2019lvis} as the boundary annotations in LVIS are much more precise, as shown in Fig.~\ref{fig:eval} (right).

MS COCO~\cite{lin2014microsoft} contains 118K images for training, and 5K images for evaluation with around 1.5M object instances and 80 categories. The annotation contains bounding box, category labels, and keypoint-based mask polygons. All instances in the dataset are exhaustively annotated.

LVIS~\cite{gupta2019lvis} contains 164K images and 2.2M high-quality instance segmentation masks for over 1000 entry-level object categories. Its images are a subset of the images from MS COCO. We keep all the annotated instances that overlap with MS COCO and re-label the categories in the same way as COCO for evaluation.

\vspace{-10pt}
\subsubsection{Evaluation Metrics.}\label{sec:eval}
As well-established problems, both semantic aware \cite{yu2017casenet,yu2018simultaneous} and agnostic \cite{arbelaez2010contour,martin2004learning} edge detection have
standard evaluation pipelines. We use the same standard ODS (optimal dataset scale) and OIS (optimal image scale) metrics to evaluate instance edge detection.

\begin{figure*}[t]
  \centering
  \includegraphics[width=1.\textwidth]{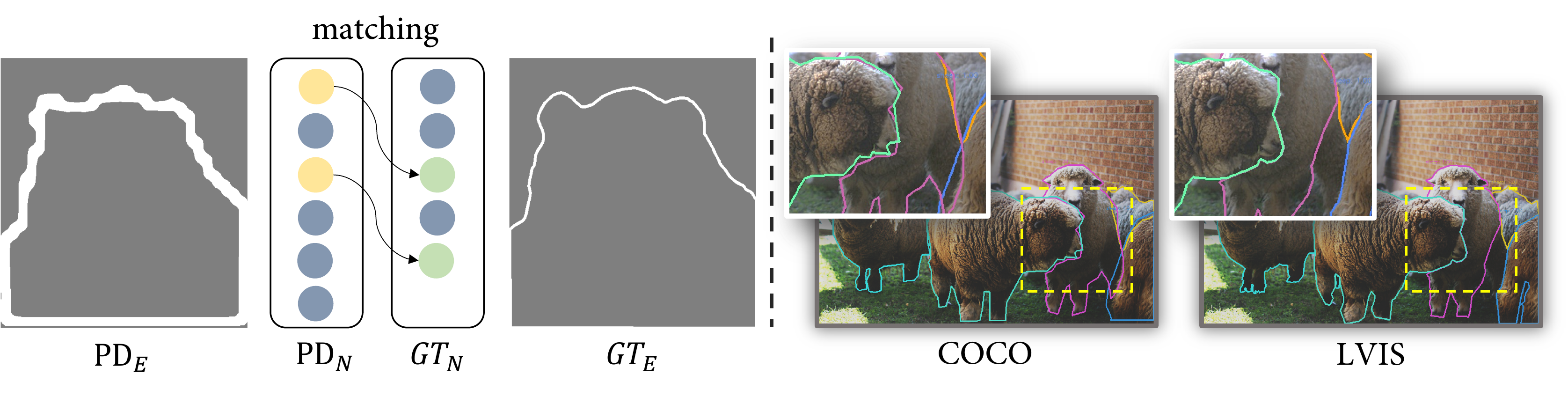}
  \caption{(Left) Bipartite matching between edge detections and ground-truth annotated edges. (Right) Annotations from COCO vs.~LVIS. It clearly shows that LVIS has more fine-grained boundary annotations than COCO.}
  \label{fig:eval}
\end{figure*}

Briefly, an edge thinning step is typically applied to produce (near) pixel-wide edges. Then, bipartite matching is used to match the predicted edges $PD$ with the ground-truth edges $GT$ (see Fig.~\ref{fig:eval} left).  Candidate matches are those whose distance is within a small pre-defined distance proportional to the image size. Then, precision $p$ and recall $r$ are computed, where precision measures the number of predicted edge points that are matched to a ground truth edge, and recall measures the number of ground truth edge points that are matched to a predicted edge.  The F-measure is then computed as $2 \cdot p \cdot r/ (p + r)$.  ODS is the best F-measure using the global optimal threshold across the entire validation set. OIS is the aggregate F-measure when the optimal threshold is chosen for each image. (We provide more details in the supplementary document.)

In addition, a recent paper~\cite{cheng2021boundary} proposes the `Boundary IoU' to supplement mask mAP for evaluating the boundary of instance segmentation. However, we argue that this is an imprecise measurement on edge, as it blurs the boundary and computes the IoU between the thick boundary and ground truth edges.

\vspace{-10pt}
\subsubsection{Implementation Details.}

\emph{Training:} We adopt the DETR framework proposed in~\cite{carion2020end} and replace the transformer decoder with the conditional decoder from Conditional DETR~\cite{meng2021conditional} to accelerate model training by $\sim$6x. For MS COCO, we train all models on 4 NVIDIA 3090 Ti GPUs with per GPU batch size of 2. For the experiments in Table~\ref{table:coco}, we use the multi-scale transformer decoder with 6 consecutive layers. For each layer, the input feature resolution is $[1/32,1/32,1/32,1/32,1/16,1/8]$ corresponding to the original image size. The dense prediction head takes in the last layer output of the FPN network with a size of 1/4 of the original resolution. In Table \ref{table:type_gt}, \ref{table:fea_res}, to reduce computation cost, we use single scale transformer decoder with a $1/32$ input resolution. And the dense prediction head takes in FPN features with $1/8$, $1/4$ of the original image resolution respectively. Most of the hyperparameters follow the implementation in~\cite{meng2021conditional}.

\begin{table*}[t]
\centering
\resizebox{0.98\textwidth}{!}{
\begin{tabular}{c|ccc|ccccc|cc}
\hline
                   &                &              &               & \multicolumn{5}{c|}{COCO}                                                                 & \multicolumn{2}{c}{LVIS}      \\ \hline
                   & {\ul \textit{Backbone}} & {\ul \textit{Epochs}} & {\ul \textit{\#Param}} & {\ul \textit{ODS}}     & {\ul \textit{OIS}}     & {\ul \textit{$AP ^{bdry}$}} & \textit{{\ul $AP^{box}$}} & {\ul \textit{$AP^{mask}$}} & {\ul \textit{ODS}}     & {\ul \textit{OIS}}     \\
Mask R-CNN         & R-50           & 50           & 44M           & 62.9          & 62.9          & {\ul 23.2}               & 40.9             & \textbf{37.0}              & 63.8          & 64.3          \\
BMask R-CNN        & R-59           & 12           & 47 M          & 44.1          & 47.3          & \textbf{23.5}      & 38.6             & 36.6              & 45.5          & 46.1          \\
BMask R-CNN        & R-50           & 50           & 47M           & 44.2          & 47.2          & 22.9               & 37.6             & 35.0              & 45.9          & 46.7          \\
Ours (Mask)        & R-50           & 50           & 46M           & 56.0          & 56.4          & 18.4               & {\ul 42.8}       & 34.0              & 60.0          & 60.4          \\
Ours (Edge)        & R-50           & 50           & 46M           & {\ul 63.1}    & {\ul 63.8}    & -                  & 42.6             & -                 & {\ul 66.2}    & {\ul 67.9}    \\
Ours (Edge + Mask) & R-50           & 50           & 47M           & \textbf{63.6} & \textbf{64.5} & 21.7               & \textbf{43.0}    & 35.0     & \textbf{66.6} & \textbf{68.3} \\ \hline
Mask R-CNN         & R-101          & 50           & 63M           & \textbf{63.7} & {\ul 63.7}    & 24.4               & {\ul 42.6}       & 38.3              & {\ul 64.7}    & {\ul 65.2}    \\
BMask R-CNN        & R-101          & 12           & 66M           & 44.7          & 48.2          & 24.7               & 40.6             & 38.0              & 46.5          & 47.0          \\
BMask R-CNN        & R-101          & 50           & 66M           & 44.8          & 47.7          & 24.3               & 40.0             & 36.7              & 46.7          & 46.1          \\
Ours (Edge)        & R-101          & 50           & 65M           & {\ul 63.6}    & \textbf{64.4} & -                  & \textbf{44.3}    & -                 & \textbf{67.1} & \textbf{68.7} \\ \hline

\end{tabular}
}
\vspace{5pt}
\caption{Edge detection, object detection, and instance segmentation results on MS COCO and LVIS.}
\vspace{-15pt}

\label{table:coco}
\end{table*}

\emph{Evaluation:} We evaluate our approach on three different tasks including object detection, instance segmentation, and edge detection. For object/instance detection, we follow \cite{lin2014microsoft} and use mAP metric for evaluation. For instance edge detection, we use ODS and OIS following~\cite{xie2015holistically}. The reason for not including AP for edge detection is because the boundary of instance segmentation will only have a probability range from [0.5,1], which will introduce errors in computing the AP score. To compensate for this, we use AP boundary~\cite{cheng2021boundary} to evaluate our model when there is a mask output.

\vspace{-10pt}
\subsubsection{Baselines.}

The most related work to ours that  simultaneously predicts instance mask and boundaries is BMask R-CNN~\cite{cheng2020boundary}, which learns a separate instance edge detection head in parallel with the mask and box heads in Mask R-CNN~\cite{he2017mask}.
In addition, since instance edges can be computed from instance segmentation masks, we also compare to the boundaries of the masks produced by Mask R-CNN~\cite{he2017mask}. Instance edge is derived from instance mask by applying a laplacian filter on the binary mask. This baseline is used to demonstrate that this way of computing instance edges is insufficient due to the bias in the mask segmentation objective, which rewards accurate prediction of interior pixels in the ground-truth mask more than those that are on the boundary (since they are relatively much fewer).

\vspace{-3pt}
\subsection{Quantitative Results}\label{sec:quan}
\vspace{-1pt}

In Table~\ref{table:coco}, we compare our approach with various state-of-the-art baselines for edge detection, object detection, and instance segmentation tasks using the COCO and LVIS datasets. For BMask R-CNN, we use the authors' publicly available codebase. For Mask R-CNN, we use Detectron2~\cite{wu2019detectron2} to train and evaluate the baseline model. For all the baseline methods, we re-train the model with 50 epochs schedule. Despite multiple attempts, we could not get BMask R-CNN trained with 50 epochs to outperform its 12 epochs model, which is why we report both of them in the table.

\begin{figure*}[t!]
  \centering
  \includegraphics[width=0.9\textwidth]{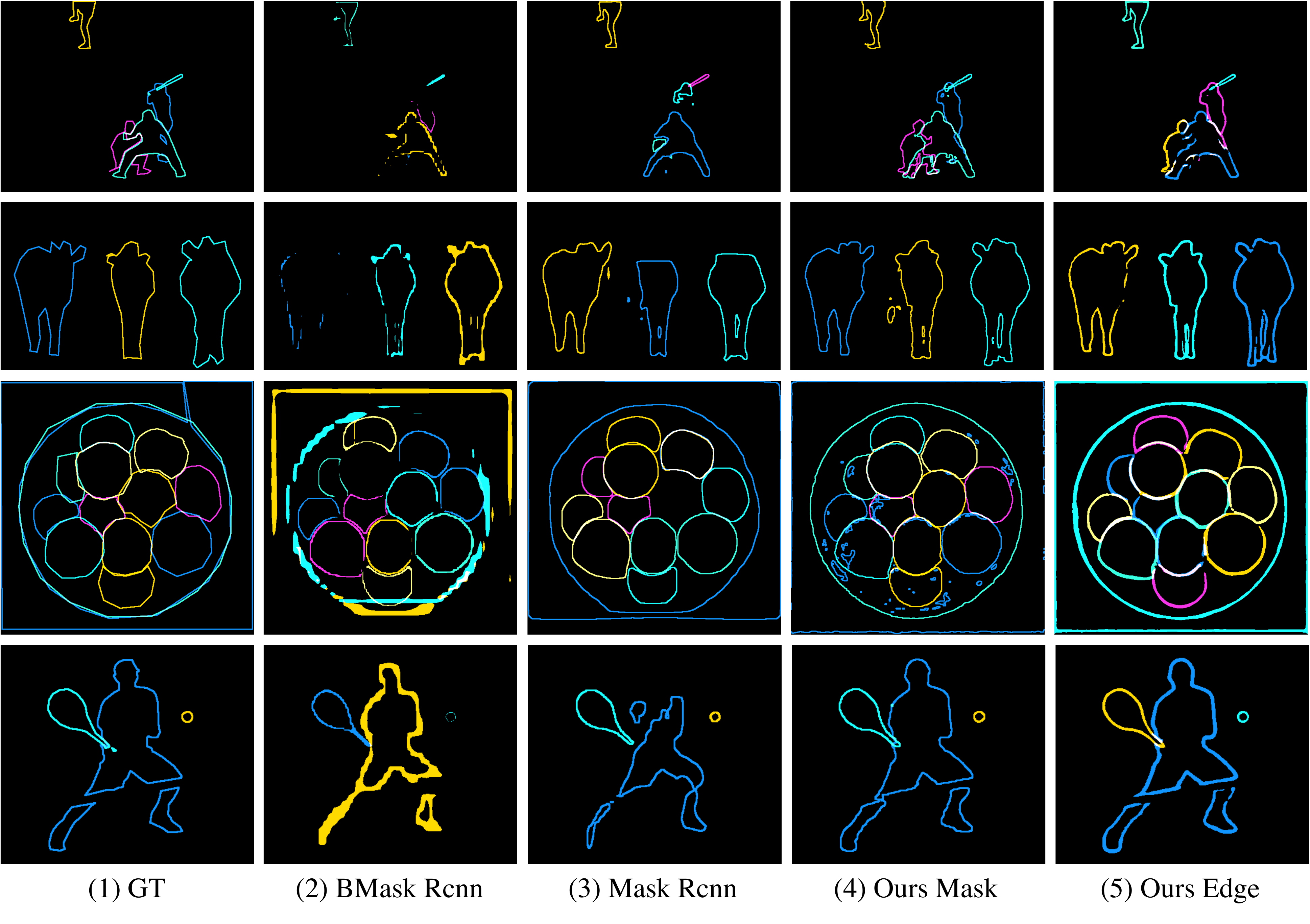}
  \vspace{-10pt}
  \caption{Qualitative results comparing to the baselines. The thickness of predicted edges varies largely for BMask R-CNN. In addition, the mask boundary quality of Mask R-CNN is not as good as that of Ours Edge and Ours Mask. Finally, there exist redundant predictions or holes in our mask variant (Ours Mask) predictions.}
  \label{fig:qua}
  \vspace{-10pt}
\end{figure*}

\vspace{-10pt}
\paragraph{Edge detection}
On the COCO dataset, our approach achieves the best results under ODS/OIS edge detection metrics compared to BMask R-CNN and Mask R-CNN. Surprisingly, we achieve $\sim$18\% better performance than BMask R-CNN, which is our closest baseline. When taking a closer look at the qualitative results in Fig.~\ref{fig:qua}, the reason becomes clear. For example, in the second column of Fig.~\ref{fig:qua}, using the same edge probability threshold, the thickness of the predicted instance boundaries for BMask R-CNN varies widely. This indicate that the model lacks a unified treatment for all instances, and thus it is harder to find a single threshold that works well for all instances in all images. The quantitative results on OIS and ODS again prove this hypothesis: the OIS improves by around 3-4 points for BMask R-CNN while it does not change a lot for all other models. In addition, because we are directly thresholding the predicted masks for Mask R-CNN and our mask variant (Ours Mask) to obtain edge detections, their OIS and ODS remain nearly constant under all mask settings. 
Apart from our better performance compared to the edge detection method of BMask R-CNN, our approach also performs better than instance segmentation methods (especially on lvis dataset with accurate ground truth): Mask R-CNN and our mask variant (Ours Mask). This is mainly due to two reasons: (1) The baseline mask predictions are inaccurate along boundaries. (2) The baseline mask can have holes inside. These observations are further illustrated in Sec.~\ref{sec:qua}. 

On the LVIS dataset, the results are consistent with those on the COCO dataset. However, in general all methods achieve better results using LVIS annotations. And our approach performs extremely well on LVIS dataset in comparison with Mask R-CNN with $\sim$ 3-4 higher on both scale of models (R-50, R-101). This is also explainable if we take a look at Fig.~\ref{fig:eval}: LVIS has more precise boundary annotations than COCO.  The predictions are usually aligned better with these more accurate annotations.

\vspace{-5pt}
\paragraph{Object detection}
Our approach also achieves the best result on box mAP with $\sim$2-4 points higher on ResNet 50 backbone with 50 epochs compared to both Mask R-CNN and BMask R-CNN. When training with a larger ResNet 101 backbone, this improvement also holds with a consistent performance gain of $\sim$2-4 points.

\vspace{-10pt}
\paragraph{Instance segmentation}
Finally, we compare with the baselines on the instance segmentation task using our model with the dense prediction head plus mask loss. It performs $\sim$2 points worse than Mask R-CNN and BMask R-CNN. One hypothesis is that training an object query containing both mask and box information has a divergent effect; e.g., object query for box detection should have the ability to locate the extreme points of an object, whereas instance segmentation requires the query to focus on the full object.

\vspace{-5pt}
\subsection{Ablation Study}
\vspace{-1pt}

\begin{table}[t]

\begin{minipage}{.5\linewidth}
\hspace{20pt}
\resizebox{0.87\textwidth}{!}{
\begin{tabular}{c|ccc|cc}
\hline
           & \multicolumn{3}{c|}{COCO}                                                            & \multicolumn{2}{c}{LVIS}                \\ \hline
           & {\ul \textit{ODS}} & {\ul \textit{OIS}} & {\ul \textit{$AP^{box}$}} & {\ul \textit{ODS}} & {\ul \textit{OIS}} \\
contour GT & 59.0               & 59.3               & 41.0                                       & 62.5               & 63.1               \\
point GT   & \textbf{63.0}      & \textbf{63.7}      & \textbf{41.5}                              & \textbf{66.7}      & \textbf{67.9}      \\ \hline
\end{tabular}}
\end{minipage}
\begin{minipage}{.5\linewidth}
\hspace{10pt}
\resizebox{0.75\textwidth}{!}{
\begin{tabular}{c|ccc|cc}
\hline
                     & \multicolumn{3}{c|}{COCO}                                           & \multicolumn{2}{c}{LVIS}                \\ \hline
{\ul \textit{ratio}} & {\ul \textit{ODS}} & {\ul \textit{OIS}} & {\ul \textit{$AP^{box}$}} & {\ul \textit{ODS}} & {\ul \textit{OIS}} \\
1/2                  & 58.4               & 59.4               & 41.3                      & 61.7               & 63.9               \\
2/3                  & 61.7               & 62.5               & 41.1                      & 64.7               & 66.4               \\
1/1                  & 63.0               & 63.7               & 41.5                      & 66.7               & 67.9               \\ \hline
\end{tabular}}
\end{minipage}
\vspace{3pt}
\caption{(Left) Varying the type of ground truth training target for edge detection. (Right)  Varying the number of ground truth edge points used for training.  With ratio 1, the
average number of keypoints across all instances is 23.}
\label{table:type_gt}
\end{table}

\paragraph{Type of ground truth.}
We first study the effect of our point supervised training objective, which models the uncertainty in the edges that are not labeled by the keypoints by assigning them a softer target score.  We compare to the training objective used in BMask R-CNN, which simply connects the keypoints to create ground-truth edges, and applies both a weighted binary cross-entropy loss and the dice loss~\cite{milletari2016v}.  As shown in Table~\ref{table:type_gt} (left), training with our point supervision objective (point) produces significantly better edge detection performance on both COCO and LVIS datasets compared to the baseline (contour). Furthermore, the improvements on edge detection also lead to a 0.5 improvement in box mAP, which demonstrates their complementary relationship.

\vspace{-10pt}
\paragraph{Number of edge points.}
A key advantage of training an edge detector with point supervision is the large reduction in annotation effort that is required.  We therefore investigate how the number of annotated edge points affects instance edge detection performance. Specifically, we sample the number of end points that are used for training from 1/1 to 2/3 to 1/2 of the full original set of annotated keypoints.  As shown in Table~\ref{table:type_gt} (right), by decreasing the annotation by 1/3 and 1/2, both ODS and OIS decreases as expected but not by a large amount.

\begin{wraptable}{tr}{0.5\textwidth}
\vspace{-15pt}
\resizebox{0.5\textwidth}{!}{
\begin{tabular}{ccc|cccc}
\hline
\multicolumn{3}{c|}{}                                                             & \multicolumn{4}{c}{COCO}                                                                 \\ \hline
{\ul \textit{box}}        & {\ul \textit{mask}}       & {\ul \textit{edge}}       & {\ul \textit{$AP^{box}$}} & {\ul \textit{$AP^{mask}$}} & {\ul \textit{ODS}} & {\ul \textit{OIS}} \\
\checkmark &                           &                           & 40.9                  & -                      & -                  & -                  \\
\checkmark & \checkmark &                           & 41.1                  & 34.5                   & 56.3               & 56.4               \\
\checkmark &                           & \checkmark & 41.3                  & -                      & 63.4               & 64.2               \\
\checkmark & \checkmark & \checkmark & \textbf{41.6}         & \textbf{35.1}          & \textbf{63.6}      & \textbf{64.5}      \\ \hline

\end{tabular}}
\caption{Varying annotation types (bounding boxes, masks, and edges) for model training.}
\vspace{-15pt}
\label{table:fea_res}
\end{wraptable}

\vspace{-10pt}
\paragraph{Annotation types.}
We also ablate our DETR based dense prediction framework under different types of annotations (box, mask, and edge). As shown in Table~\ref{table:fea_res}, by adding mask and edge objectives, box prediction improves by 0.2 and 0.4 points respectively. We can also conclude from the table that training with an edge objective leads to a much better edge detection result in comparison with training with a mask objective. This further proves our argument that instance edge detection is different from instance segmentation. Last but not least, by simply adding an edge objective to mask objective, AP box further improves by 0.3 points, and AP mask improves by a good margin with 0.6 point.

\vspace{-10pt}
\subsection{Qualitative Results}\label{sec:qua}
For a fair comparison, all qualitative results use the models trained with a ResNet 50 backbone with 1x schedule. We threshold the mask probability with $0.5$ to obtain the binary mask together with their boundaries. And for edge detection methods, we use 0.7 as a threshold to filter out noisy predictions. As stated in Sec.~\ref{sec:quan}, we can observe clear reasons for why our approach achieves better performance for edge detection. For example, in the second row of Fig.~\ref{fig:qua}, while the blue cow predicted by BMask R-CNN is nearly thresholded out, the edge of the yellow cow remains very thick. This is the primary reason for BMask R-CNN's low performance. Further, the third column of Fig.~\ref{fig:qua} shows the results of Mask R-CNN, which clearly indicate that it is usually unable to predict the boundaries well (e.g. the blue cow in second row, the person in fourth row) compared with our mask and edge models. Last but not least, although our mask variant (Ours Mask) usually generates high quality boundaries, when the mask is large, redundant predictions or holes can appear in the mask as shown in the fourth column.

\vspace{-5pt}
\section{Conclusion and Limitations}
\vspace{-2pt}

We introduced a novel point supervised transformer model for edge detection. In an extension to the DETR object detector, we introduce a multi-scale transformer decoder and a dense prediction head that could be easily applied to both instance segmentation and edge detection. Although our approach achieves good results for object and edge detection, it does not perform as well on instance segmentation. This is likely because of the divergent objective function -- for the same object query, instance segmentation requires focusing on the whole object but edge/object detection requires focusing more on object boundaries.

\appendix
\section{Appendix}
\subsection{Comparison with edge detection methods}
In the main paper, we  compared with strong object detection based baseline methods that perform instance segmentation (Mask RCNN) and instance edge detection (BMask RCNN). In this section, we compare our method with state-of-the-art edge detection methods:  RINDNet~\cite{pu2021rindnet}, CaseNet~\cite{yu2017casenet} and HED~\cite{xie2015holistically}. In order to compare with these edge detection methods, we first use the bounding box provided by our approach (Row 4, in Table~\ref{table:edge}) to crop the image, and then apply the corresponding edge detection baseline method. We adopt the implementations from \url{https://github.com/MengyangPu/RINDNet}.  Table~\ref{table:edge}  clearly shows that our approach outperforms the specialized edge detection approaches by a large margin. This is mainly because these edge detection methods not only detect object boundaries, but also have a tendency to predict inner edges within an object. 

\begin{table}[]
\centering
\begin{tabular}{c|cccc}
\hline
\multirow{2}{*}{{\ul \textbf{Method}}} & \multicolumn{2}{c}{{\ul \textbf{COCO}}} & \multicolumn{2}{c}{{\ul \textbf{LVIS}}}             \\
                                       & ODS                & OIS                & ODS                      & OIS                      \\ \hline
HED                            &        44.2     &      46.3      & 47.7                     & 51.5                     \\
CASENet                                &    44.0        &    46.0      & 47.8                     & 51.2                     \\
RINDNet                                &    41.9       &    43.4         & 45.0                     & 47.8                     \\
Ours (Edge)                            & 63.1               & 63.8               & \multicolumn{1}{l}{66.2} & \multicolumn{1}{l}{67.9} \\
Ours (Edge + Mask)                     & \textbf{65.8}               & \textbf{64.2}               & \multicolumn{1}{l}{ \textbf{66.6}} & \multicolumn{1}{l}{\textbf{68.3}} \\ \hline
\end{tabular}
\vspace{5pt}
\caption{Comparison with specialized edge detection methods on instance edge detection.}
\label{table:edge}
\end{table}

\vspace{-35pt}

\subsection{Evaluation Metric}
\paragraph{Edge Thinning}
As the predicted contours of edge detectors are usually not a pixel-wide line, it is standard to apply an edge thinning technique before evaluation. In this work, we use morphological thinning~\cite{guo1989parallel,lam1992thinning}:
\begin{gather}
    X_{i} = (X_{i-1} \ominus F) - (X_{i-1} \ominus F) \circ F \\
    X =  \cup_{i=0}^{i=N} (X_i)
\end{gather}
where $X_0$ is the input edge and $X$ is the output thinned edge map. $F$ represents structure kernels, and $\ominus, \circ$ are erosion and dilation operation, respectively.

\paragraph{Bipartite Matching}
We apply bipartite matching between the (thinned) predicted edges and ground truth edges. As shown in Fig.~4 (Main Paper), given ground truth edge map $GT_E$ and predicted edge map $PD_E$, we first extract all the positive pixels as graph nodes denoted as $GT_N=\{g_0,g_1, ...,g_n\}$ and $PD_N=\{p_0,p_1, ...,p_n\}$. Given a pre-specified max distance $\lambda$ and image size $H \times W$, all predicted nodes within distance $d=\sqrt{H^2 + W^2}*\lambda$ of $g_i$ are regarded as the assignment candidates of $g_i$. The formal specification of the candidate set is:
\begin{equation}
    M(g_i) = \{p_k \in PD_N \ | \ D(g_i, p_k) < d \}
\end{equation}

Then, the problem becomes a minimum cost bipartite assignment problem to minimize the following cost function:
\begin{gather}
    min \sum_{i=0}^{n} C(g_i, p_k \in M(g_i))
\end{gather}
where $D(x,y)$ is the euclidean distance between $x,y$ nodes. Following~\cite{martin2004learning}, we use the assignment algorithm proposed in~\cite{goldberg1995efficient,cherkassky1997implementing} that has $O(N)$ time complexity for the sparse setting.

\paragraph{Score Accumulation} 
Given an input image $I$ with size $[h,w,3]$, we denote the predicted matched instance edge detection as $PD_{E_*}$ with size $[m,h,w]$ and ground truth instance edges as $GT_{E_*}$ with size $[m,h,w]$, where $m$ is the number of instances in image $I$. We apply bipartite matching separately on each instance and accumulate results image-wise. As shown in Fig.~4 (Main Paper), the output of bipartite matching contains, for each instance $i$: (1) the true positive pixels of predicted instance edge map $PD_{E_i}^{tp}$ with size $[1,h,w]$; (2) the true positive pixels of ground truth instance edge map $GT_{E_i}^{tp}$ with size $[1,h,w]$. The precision and recall for a single image is calculated as:
\begin{gather}
    p = \sum_{i=0}^{m} PD_{E_i}^{tp} / \sum_{i=0}^{m} PD_{E_i}; \
    r = \sum_{i=0}^{m} GT_{E_i}^{tp} / \sum_{i=0}^{m} GT_{E_i} 
\end{gather}
where $p$ and $r$ represents precision and recall respectively. The final scores are averaged across images in the validation set. 

As the output of the edge detector is usually probabilities, a threshold is needed to obtain a binary edge prediction.  In practice, multiple thresholds are applied during validation. In this paper, we use thresholds in $20$ intervals in range $[0,1)$. Following~\cite{xie2015holistically}, we compute ODS and OIS to measure edge prediction quality. ODS uses the threshold that maximizes the F-score ($2pr/(p+r)$) across the validation set, whereas OIS maximizes per image F-score. For each threshold, we can plot a point on the precision/recall curve.

\subsection{Qualitative Results}
\begin{figure*}[t]
  \centering
  \includegraphics[width=0.95\textwidth]{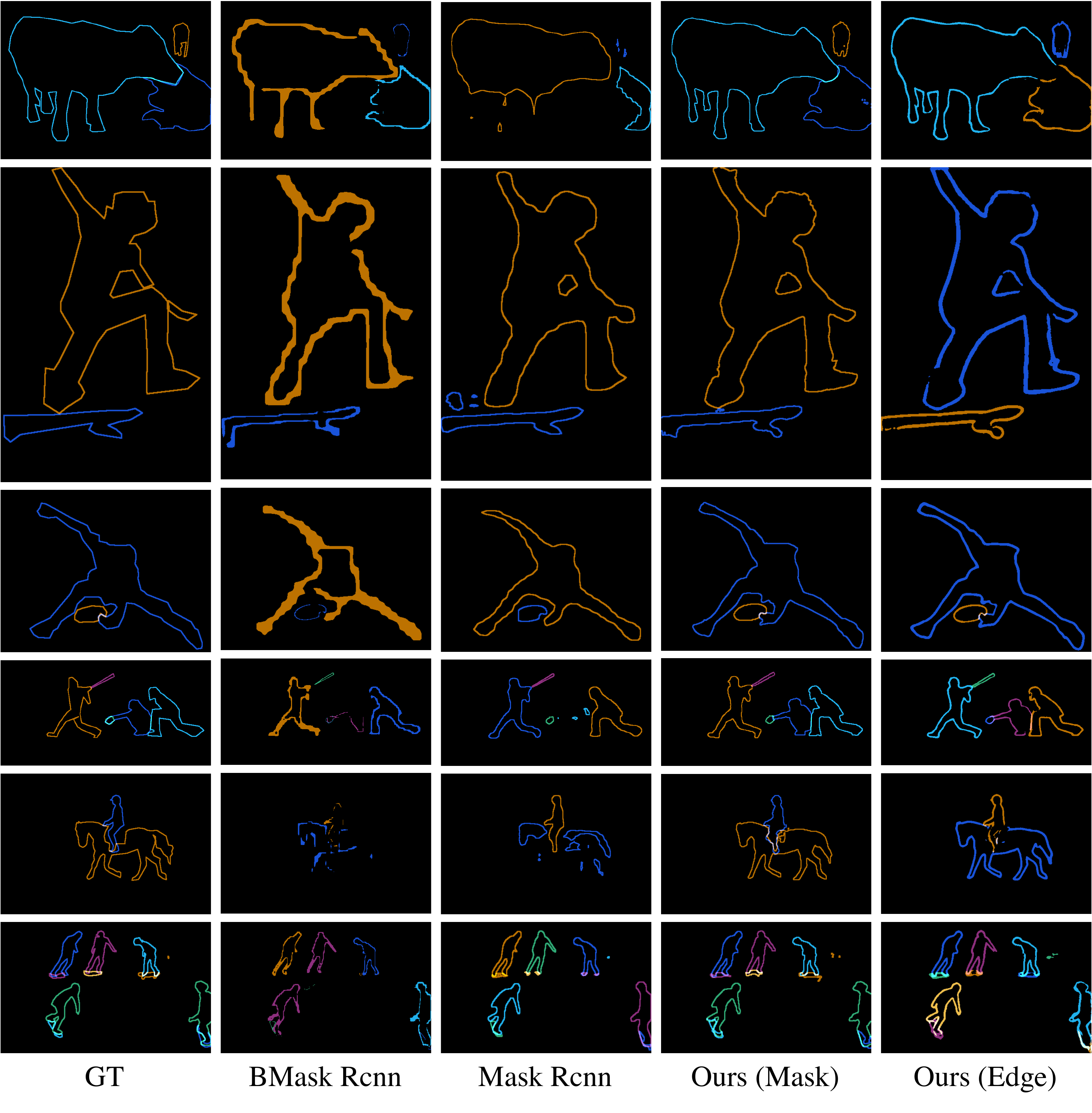}
  \caption{Additional qualitative results.}
  \label{fig:qua1}
\end{figure*}

\begin{figure*}[t]
  \centering
  \includegraphics[width=0.95\textwidth]{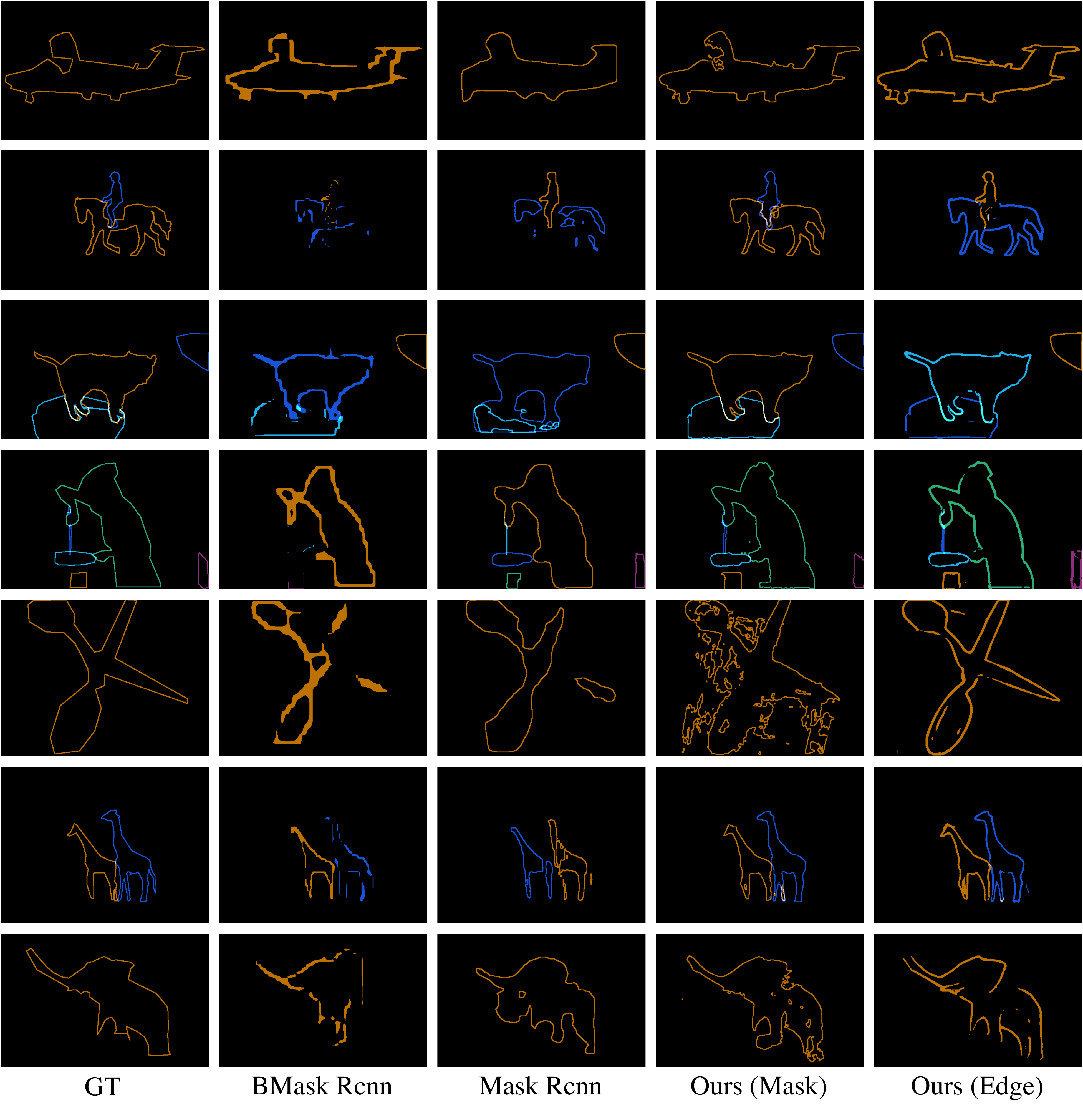}
  \caption{Additional qualitative Results}
  \label{fig:qua2}
\end{figure*}

In Figures~\ref{fig:qua1} and~\ref{fig:qua2}, we show additional qualitative edge detection results in comparison to the BMask R-CNN and Mask R-CNN baselines.

\clearpage
\bibliographystyle{splncs04}
\bibliography{egbib}
\end{document}